\documentclass{article}

\usepackage{microtype}
\usepackage{graphicx}
\usepackage{subfigure}
\usepackage{booktabs} 
\usepackage[utf8]{inputenc}

\usepackage{hyperref}

\usepackage[accepted]{icml2023}

\usepackage{amsmath}
\usepackage{amssymb}
\usepackage{mathtools}
\usepackage{amsthm}

\usepackage[capitalize,noabbrev]{cleveref}

\theoremstyle{plain}

\theoremstyle{definition}

\theoremstyle{remark}

\usepackage[textsize=tiny]{todonotes}

\DeclareSymbolFont{extraup}{U}{zavm}{m}{n}
\DeclareMathSymbol{\vardiamond}{\mathalpha}{extraup}{87}

\icmltitlerunning{Chain-of-Thought Hub}

\begin{document}

\twocolumn[
\icmltitle{Chain-of-Thought Hub: A Continuous Effort to\\
Measure Large Language Models' Reasoning Performance}



\icmlsetsymbol{equal}{*}

\begin{icmlauthorlist}
\icmlauthor{Yao Fu$^{\spadesuit}$}{}
\icmlauthor{Litu Ou$^{\spadesuit}$}{}
\icmlauthor{Mingyu Chen$^{\spadesuit}$}{}
\icmlauthor{Yuhao Wan$^\vardiamond$}{}
\icmlauthor{Hao Peng$^{\clubsuit}$}{}
\icmlauthor{Tushar Khot$^{\clubsuit}$}{}
\end{icmlauthorlist}

\center{$^{\spadesuit}$University of Edinburgh\quad\quad 
$^\vardiamond$University of Washington\quad\quad$^{\clubsuit}$Allen Institute for AI}
\center{\{yao.fu, s1970716, s2331360\}@ed.ac.uk\quad\quad yuhaowan@cs.washington.edu\quad\quad\{haop, tushark\}@allenai.org}
\center{\url{https://github.com/FranxYao/chain-of-thought-hub}}


\icmlcorrespondingauthor{Yao Fu}{yao.fu@ed.ac.uk}

\icmlkeywords{Machine Learning, ICML}

\vskip 0.3in
]




\begin{abstract}
As large language models (LLMs) are continuously being developed, their evaluation becomes increasingly important yet challenging. 
This work proposes Chain-of-Thought Hub, an open-source evaluation suite on the multi-step reasoning capabilities of large language models. 
We are interested in this setting for two reasons:
(1) from the behavior of GPT and PaLM model family, 
we observe that complex reasoning is likely to be a key differentiator between weaker and stronger LLMs;
(2) we envisage large language models to become the next-generation computational platform and foster an ecosystem of LLM-based new applications, this naturally requires the foundation models to perform complex tasks that often involve the composition of linguistic and logical operations. 
Our approach is to compile a suite of challenging reasoning benchmarks to  track the progress of LLMs. 
Our current results show that: 
(1) model scale clearly correlates with reasoning capabilities;
(2) As of May 2023, Claude-v1.3 and PaLM-2 are the only two models that are comparable with GPT-4, while open-sourced models still lag behind;
(3) LLaMA-65B performs closely to code-davinci-002, indicating that with successful further development such as reinforcement learning from human feedback (RLHF), it has great potential to be close to GPT-3.5-Turbo. 
Our results also suggest that for the open-source efforts to catch up, the community may focus more on building better base models and exploring RLHF.
\end{abstract}

\section{Introduction}
\label{sec:intro}

Recently, the field of AI has been significantly impressed by the advances in large language models. 
LLMs exhibit multi-dimensional capabilities, and their evaluation is challenging. 
Generally, tuning a base language model into a chatbot is relatively easy, as demonstrated by the large variety of LLaMA-based~\citep{touvron2023llama} models like Alpaca~\citep{alpaca}, Vicuna~\citep{vicuna2023}, Koala~\citep{koala_blogpost_2023}, Dolly~\citep{dolly}, and so on. 
In a chitchat, all these models may perform superficially similarly to GPT-3.5-Turbo~\citep{arnav2023falsepromise}. 
At the current stage, the community is eager to know what are 
the key factors that clearly differentiate the 
better-performing models from the underperforming ones.

In this work, we consider the evaluation of complex reasoning.
As noted by \citet{openai-blog-gpt4}, ``In a casual conversation, the distinction between GPT-3.5 and GPT-4 can be subtle. The difference comes out when \textit{the complexity of the task reaches a sufficient threshold}''.
A similar observation is made by the Google PaLM model family, as their developers discover that large models' chain-of-thought reasoning capability is clearly stronger than smaller models~\citep{wei2022chain, wei2022emergent}. 
These observations indicate that the ability to perform complex tasks
is a key metric.

The capability of performing complex reasoning is 
crucial for the models to become the next-generation computation platform. 
One example initiative is LangChain\footnote{https://github.com/hwchase17/langchain} where developers build applications powered by backend LLM engines, which generally require the model to perform complex tasks. 
Here the vision of pushing LLMs as the foundation of a new computational ecosystem also serves as a strong motivation to measure the models' reasoning performance.

To incentivize the research efforts in improving language models' reasoning performance, we propose the chain-of-thought hub (CoT Hub), a continuous open-source effort that tracks LLMs' reasoning capability using a carefully curated evaluation suite. 
CoT Hub is the first comprehensive comparison of very large LMs on reasoning benchmarks and
currently consists of 19 major language models' (including the GPT, Claude, PaLM and LLaMA) performance on 6 benchmarks and more than 100 subtasks (including bi-lingual reasoning capabilities in Chinese), 
and we are continuously adding new models and datasets.

Observations made in CoT Hub shed light on many insights into LLM development:
(1) the reasoning performance of LLMs highly correlates with models' scales;
(2) as of May 2023, PaLM and Claude\footnote{https://www.anthropic.com/index/introducing-claude} are the only two model families that are comparable to (yet slightly worse than) the GPT model family;
(2) LLaMA 65B~\cite{touvron2023llama} 
the strongest open LLM to date,
performs closely to code-davinci-002, the base model of GPT-3.5 family\footnote{https://platform.openai.com/docs/model-index-for-researchers}. 
This indicates that \textbf{if aligned properly} (by doing supervised finetuning (SFT) and reinforcement learning from human feedback (RLHF) right) 
\textbf{LLaMA 65B can potentially improve further and perform on par with ChatGPT-3.5}. 
We hope our work gives meaningful guidance to the community's development of deployable LLMs.

\section{Method}
\label{sec:intro}
In this section we discuss the construction of Chain-of-Thought Hub. We first discuss our method for test data collection,
then we discuss how we obtain the model performance on our test suite. 
Our main goal is to curate a high-quality collection of datasets that (1) is closely related to the actual usage of LLMs; (2) clearly differentiate the performance of stronger and weaker language models. 
We consider the following datasets:

\begin{description}
\item [GSM8k] A widely used math reasoning datasets consisting of 8k problems that jointly test models' ability of arithmetic reasoning and composing math steps using language~\citep{cobbe2021training}.
\item [MATH] A suite of challenging datasets consisting of 12k problems within 7 categories testing the models' advanced math and science reasoning. The problems in this dataset are very hard because they come from mathematics competitions written in Latex. Even GPT-4 has only 42.5\% performance
~\citep{hendrycksmath2021}. 
\item [MMLU] An evaluation suite of 15k problems within 57 subjects testing model's high-school and college-level knowledge and reasoning~\citep{hendrycks2020measuring}.
\item [BigBench Hard] A suite of language and symbolic reasoning tasks consisting 6.5k problems within 23 subsets that are particularly suitable for testing chain-of-thought prompting~\citep{suzgun2022challenging}. 
\item [HumanEval] A handwritten dataset of 164 Python programming problems with text comments and docstrings testing the models' coding ability~\citep{chen2021evaluating}. 
\item [C-Eval] A Chinese evaluation suite for foundation models consisting of 13k multi-choice questions spanning 52 diverse disciplines and four difficulty levels~\citep{huang2023c}. 
\end{description}

We note that most of these datasets are already used in the evaluation of leading large language models, such as GPT-4~\citep{gpt-4} and PaLM-2~\citep{anil2023palm}. 

\textbf{Few-Shot Chain-of-thought Prompting}\quad\quad
We use few-shot chain-of-thought prompting to evaluate LLMs. 
This marks a clear difference between our evaluation and the majority of other concurrent evaluations like HeLM~\citep{liang2022holistic}, as most of them use answer-only prompting. 
We also emphasize that we use few-shot, rather than zero-shot prompting, because few-shot is a capability that exist in both pretrained and instruction-tuned checkpoints, v.s., zero-shot is more suitable for instruction-tuned checkpoints and may under-estimate the pretrained checkpoints. 

\textbf{Comparison to existing and concurrent work}\quad\quad
There are many great existing evaluation suites for large language models, such as HeLM, Chatbot Arena\footnote{https://leaderboard.lmsys.org/}, and Open LLM Leaderboard\footnote{https://huggingface.co/spaces/HuggingFaceH4/open\_llm\_leaderboard}. 
The major difference between this work and these works are:
(1) HeLM evaluates a significantly wider spectrum of tasks while we focus on evaluating reasoning. 
Most of the results from this work use chain-of-thought prompting (hence the name ``Chain-of-Thought Hub'') whereas HeLM mainly uses answer-only prompting (without CoT). 
(2) Chatbot Arena evaluate the dialog user preference we evaluate reasoning. 
(3) Open LLM Leaderboard focus on open-sourced LLMs, we jointly consider major LLMs, either open-sourced or not. 

\textbf{Using final answer accuracy as a proxy for reasoning capability}\quad\quad
Most of the datasets we consider share one pattern: to reach the final answer (either a number for math problems, a choice for multi-choice problems, or a fixed output for coding), the model needs to figure out the intermediate steps toward that answer. 
When evaluating, we only use the final answer accuracy but do not consider the correctness of intermediate steps. 
This is because empirically, the correctness of intermediate steps is strongly correlated with the final accuracy. 
If the intermediate steps are very wrong, the model is less likely to reach the final answer. 
If the final answer is correct, the intermediate steps are generally good enough~\citep{wei2022chain, lewkowycz2022solving}.


\section{Experiments}
\label{sec:results}
\begin{table*}[ht!]
  \caption{
  \textbf{Overall model performance on Chain-of-Thought Hub}. Numbers with an asterisk* are from our test scripts.
  For model types, base means the model checkpoint after pretraining, SIFT means supervised instruction finetuning. 
  Others are from their corresponding papers. 
  We observe: (1) there exist a gap between leading LLMs (GPT, Claude and PaLM) and open-source (LLaMA and FlanT5);
  (2) most leading LLMs are after RLHF, indicating the  opportunity of improving open-sourced models using this technique;
  (3). model performance is generally correlated with model scale, indicating further opportunities in scaling, especially for open-source models.
  We further highlight that among open-sourced models, LLaMA 65B performs close to code-davinci-002, the base model of ChatGPT. 
  This suggests that if RLHF is done right on LLaMA 65B, it may become close to ChatGPT. 
  }
  \label{tab:exp:overall}
  \centering
  \begin{tabular}{@{}lll|cccccc@{}}
      \toprule
       Model & \#Params & Type & GSM8k & MATH & MMLU & BBH & HumanEval & C-Eval\\ 
       \midrule
      GPT-4 & ? & RLHF & 92.0 & 42.5 & 86.4 & - & 67.0 & 68.7*\\ 
      claude-v1.3& ? & RLHF & 81.8* & - & 74.8*& 67.3* & - & 54.2* \\ 
      PaLM-2	& ?	& Base	& 80.7	& 34.3	& 78.3	& 78.1 & - & -\\ 
      gpt-3.5-turbo& ? & RLHF & 74.9*	& -	& 67.3* & 	70.1* & 48.1 & 54.4*\\
      claude-instant-v1.0 &	? &	RLHF&	70.8*&	-&	-&	66.9* & - & 54.9*\\ 
      text-davinci-003 &	?&	RLHF&	-&	-&	64.6&	70.7 & - & - \\ 
      code-davinci-002 &	?&	Base&	66.6&	19.1&	64.5 & 73.7 & 47.0 & -\\
      Minerva	& 540B &	SIFT&	58.8&	33.6&	-&	- & - & -\\ 
      Flan-PaLM &	540B &	SIFT & 	- &	- &	70.9 &	66.3 & - & -\\
      Flan-U-PaLM & 	540B &	SIFT & 	- &	- &	69.8 & 	64.9 & - & -\\ 
      PaLM &	540B &	Base &	56.9 &	8.8 &	62.9 &	62.0 & 26.2 & -\\
      text-davinci-002 &	? &	SIFT &	55.4 &	- &	60.0 & 	67.2 & - & -\\
      PaLM &	64B &	Base &	52.4 &	4.4 &	49.0 &	42.3 & - & -\\ 
      LLaMA &	65B &	Base &	50.9 &	10.6 &	63.4 &	- & 23.7 & 38.8* \\ 
      LLaMA &	33B &	Base &	35.6 &	7.1 &	57.8 &	- & 21.7 & -\\
      LLaMA &	13B &	Base &	17.8 &	3.9 &	46.9 &	- & 15.8 & -\\ 
      Flan-T5	& 11B &	SIFT &	16.1* &	- &	48.6 &	41.4 & - & -\\ 
      LLaMA &	7B &	Base &	11.0 &	2.9 & 35.1 &	- & 10.5 & -\\
      Flan-T5 &	3B &	SIFT &	13.5* &	- &	45.5 &	35.2 & - & -\\ 
      \bottomrule
  \end{tabular}
\end{table*}
\begin{figure*}
  \centering
  \includegraphics[width=\linewidth]{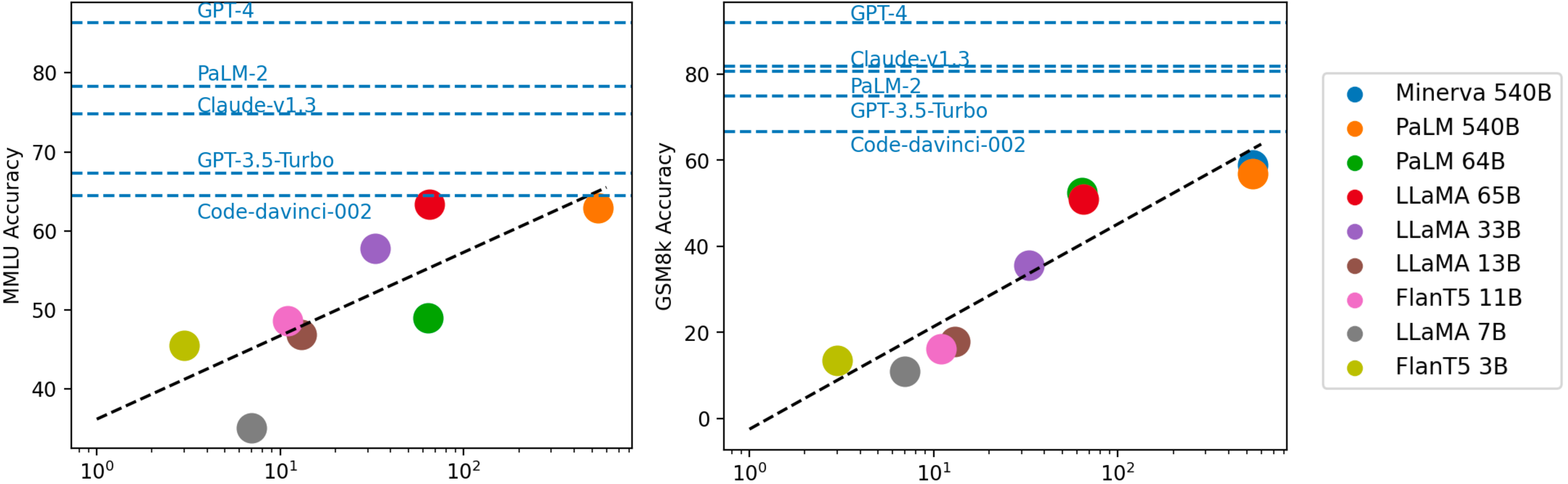}
  \caption{\label{fig:ranking} 
  X-axis means the log of the model scale measured in billion parameters. 
  We observe that model performance is generally correlated with scale, approximately showing a log-linear trend.
  Models without disclosing their scale generally perform better than models disclosing scale information. 
  Our observations also indicate that the open-source community may still needs to explore/ figure out ``the moat'' about the scaling and RLHF for further improvements.
  }
\end{figure*}

First we discuss the model families we consider. 
We focus on the popular models in production, including GPT, Claude, PaLM, LLaMA, and T5 model families,  specifically: 

\begin{description}
    \item [OpenAI GPT] including GPT-4 (currently strongest), GPT-3.5-Turbo (faster but less capable), text-davinci-003, text-davinci-002, and code-davinci-002 (important previous versions before Turbo). See~\citet{fu2022gptroadmap} for a comprehensive discussion. 
    \item [Anthropic Claude] including claude-v1.3 (slower but more capable) and claude-instant-v1.0 (faster but less capable)\footnote{https://console.anthropic.com/docs/api/reference}. 
    Strong competitor's GPT models. 
    \item [Google PaLM] including PaLM, PaLM-2, and their instruction-tuned versions (FLan-PaLM and Flan-U-PaLM). Strong base and instruction-tuned models. 
    \item [Meta LLaMA] including the 7B, 13B, 33B and 65B variants. Important open-sourced base models. 
    \item [Google FlanT5] instruction-tuned T5 models demonstrating strong performance in the smaller model regime. 
\end{description}

We report these models' performance on our CoT Hub suite. 
We note that due to the wide spectrum of the tasks and models we consider, the evaluation is nontrivial and even running inference takes effort. 
In addition, there are models that do not offer public access (such as PaLM), such that evaluating them is difficult. 
For these reasons, we report numbers using the following strategy:
if the performance of a model is already reported in a paper, we refer to that paper;
otherwise, we test them by ourselves. 
Note that this strategy is not comprehensive, as we still have a fraction of untested non-public models on some datasets. 
This is partly the reason we view our CoT Hub as a continuous effort. 

Table~\ref{tab:exp:overall} shows the overall results.
We rank the models using GSM8k performance because it is a classical benchmark testing models' reasoning capabilities. 
Numbers marked by an asterisk are tested by ourselves, others are from the following sources:
GPT-4 and PaLM-2 results are from their technical report~\citep{gpt-4, anil2023palm} respectively;
GPT-3.5-Turbo's performance on HumanEval is also from~\citet{gpt-4}. 
Text-davinci-003, code-davinci-002 and text-davinci-002 performance are from the appendix in~\citet{chung2022scaling} and from~\citet{fu2022complexity}.
Minerva's performance is from~\citet{lewkowycz2022solving}.
PaLM's performance is from~\citet{chowdhery2022palm}. 
Flan-PaLM and FlanT5 performance are from~\citet{chung2022scaling}.
LLaMA's performance is from~\citet{touvron2023llama}.

\textbf{The gap between open-source and leading LLMs}\quad\quad
In general, we observe a performance discrepancy between open-sourced models (like LLaMA and FlanT5) and close-sourced models (GPT, Claude and PaLM).
Importantly, the performance of open-sourced models seems to be upper bounded by LLaMA 65B. 

\textbf{Leading LLMs are after RLHF}\quad\quad
We observe that except for PaLM-2, the top 6 models on the leaderboard are after reinforcement learning from human feedback. 
This strongly indicates the effectiveness of RLHF.
Given that RLHF is still an underexplored area, 
we strongly encourage the community to study more on this topic. 

\textbf{Correlation between model scale and reasoning}\quad\quad 
We further study the relationship between model scale and models' reasoning performance by visualizing model performance against model scale. 
Results are shown in Fig.~\ref{fig:ranking}. 
We see that:
(1) generally, model performance is correlated with model scale, showing approximately a log-linear trend;
(2) models that do not disclose their scale generally perform better than models that do, indicating that there is still a gap between open-source and close-source.  

\textbf{On the potential of LLaMA-65B}\quad\quad 
Finally, we would like to highlight the impressive performance of LLaMA 65B.
On MMLU it is close to code-davinci-002, the base model of GPT-3.5 series. 
On GSM8k, it is worse (presumably because it is not trained on code) but close and much better than other open-sourced models (presumably because it is trained to Chinchilla-optimal~\citealp{hoffmann2022training}).
Combining this observation with the fact that GPT-3.5-Turbo (ChatGPT) is an RLHF model based on code-davinci-002, \textbf{it may be possible to reproduce ChatGPT based on LLaMA 65B by applying the RLHF techniques} discussed in DeepMind Sparrow~\citep{glaese2022improving} and Anthropic  Claude~\citep{askell2021general, bai2022training, bai2022constitutional}.

\section{Conclusion and Future Work}
\label{sec:intro}
In this work, we propose Chain-of-Thought Hub, an open-source, continuous effort to measure the reasoning capability of very large language models. 
Our results clearly show the performance differences between smaller and larger models, and between close-source and open-source models. 

After carefully examining the results, we show two important directions for further improving open-sourced models: building better base models and exploring RLHF. 
We also point out the great potential of LLaMA 65B: if aligned properly by better SFT and RLHF, it could be possible to perform on par with ChatGPT-3.5. 

In the future, we plan to further extend CoT Hub by: 
(1) including more carefully chozen reasoning datasets, especially datasets measuring commonsense reasoning, math theorem proving, and the ability to call outside APIs;
(2) including more language models, such as LLaMA-based, instruction-finetuned models like Vicuna\footnote{https://lmsys.org/blog/2023-03-30-vicuna/} and models through API access like Cohere\footnote{https://cohere.com/generate} and PaLM-2 chat-bison-001\footnote{https://cloud.google.com/vertex-ai}.
(3) exploring methods for solving MATH, the probably most challenging datasets (recall that it consists of math-
ematics competitions written in Latex), by calling APIs that compute symbolic and numerical calculus (like Wolfram Alpha\footnote{https://www.wolframalpha.com/}). 
In summary, we believe our work serves as an evaluation platform that guides the development of open-source large language models.

\bibliography{example_paper}

\begin{thebibliography}{28}
\providecommand{\natexlab}[1]{#1}
\providecommand{\url}[1]{\texttt{#1}}
\expandafter\ifx\csname urlstyle\endcsname\relax
  \providecommand{\doi}[1]{doi: #1}\else
  \providecommand{\doi}{doi: \begingroup \urlstyle{rm}\Url}\fi

\bibitem[Anil et~al.(2023)Anil, Dai, Firat, Johnson, Lepikhin, Passos, Shakeri,
  Taropa, Bailey, Chen, et~al.]{anil2023palm}
Anil, R., Dai, A.~M., Firat, O., Johnson, M., Lepikhin, D., Passos, A.,
  Shakeri, S., Taropa, E., Bailey, P., Chen, Z., et~al.
\newblock Palm 2 technical report.
\newblock \emph{arXiv preprint arXiv:2305.10403}, 2023.

\bibitem[Askell et~al.(2021)Askell, Bai, Chen, Drain, Ganguli, Henighan, Jones,
  Joseph, Mann, DasSarma, et~al.]{askell2021general}
Askell, A., Bai, Y., Chen, A., Drain, D., Ganguli, D., Henighan, T., Jones, A.,
  Joseph, N., Mann, B., DasSarma, N., et~al.
\newblock A general language assistant as a laboratory for alignment.
\newblock \emph{arXiv preprint arXiv:2112.00861}, 2021.

\bibitem[Bai et~al.(2022{\natexlab{a}})Bai, Jones, Ndousse, Askell, Chen,
  DasSarma, Drain, Fort, Ganguli, Henighan, et~al.]{bai2022training}
Bai, Y., Jones, A., Ndousse, K., Askell, A., Chen, A., DasSarma, N., Drain, D.,
  Fort, S., Ganguli, D., Henighan, T., et~al.
\newblock Training a helpful and harmless assistant with reinforcement learning
  from human feedback.
\newblock \emph{arXiv preprint arXiv:2204.05862}, 2022{\natexlab{a}}.

\bibitem[Bai et~al.(2022{\natexlab{b}})Bai, Kadavath, Kundu, Askell, Kernion,
  Jones, Chen, Goldie, Mirhoseini, McKinnon, et~al.]{bai2022constitutional}
Bai, Y., Kadavath, S., Kundu, S., Askell, A., Kernion, J., Jones, A., Chen, A.,
  Goldie, A., Mirhoseini, A., McKinnon, C., et~al.
\newblock Constitutional ai: Harmlessness from ai feedback.
\newblock \emph{arXiv preprint arXiv:2212.08073}, 2022{\natexlab{b}}.

\bibitem[Chen et~al.(2021)Chen, Tworek, Jun, Yuan, Pinto, Kaplan, Edwards,
  Burda, Joseph, Brockman, et~al.]{chen2021evaluating}
Chen, M., Tworek, J., Jun, H., Yuan, Q., Pinto, H. P. d.~O., Kaplan, J.,
  Edwards, H., Burda, Y., Joseph, N., Brockman, G., et~al.
\newblock Evaluating large language models trained on code.
\newblock \emph{arXiv preprint arXiv:2107.03374}, 2021.

\bibitem[Chiang et~al.(2023)Chiang, Li, Lin, Sheng, Wu, Zhang, Zheng, Zhuang,
  Zhuang, Gonzalez, Stoica, and Xing]{vicuna2023}
Chiang, W.-L., Li, Z., Lin, Z., Sheng, Y., Wu, Z., Zhang, H., Zheng, L.,
  Zhuang, S., Zhuang, Y., Gonzalez, J.~E., Stoica, I., and Xing, E.~P.
\newblock Vicuna: An open-source chatbot impressing gpt-4 with 90\%* chatgpt
  quality, March 2023.
\newblock URL \url{https://lmsys.org/blog/2023-03-30-vicuna/}.

\bibitem[Chowdhery et~al.(2022)Chowdhery, Narang, Devlin, Bosma, Mishra,
  Roberts, Barham, Chung, Sutton, Gehrmann, et~al.]{chowdhery2022palm}
Chowdhery, A., Narang, S., Devlin, J., Bosma, M., Mishra, G., Roberts, A.,
  Barham, P., Chung, H.~W., Sutton, C., Gehrmann, S., et~al.
\newblock Palm: Scaling language modeling with pathways.
\newblock \emph{arXiv preprint arXiv:2204.02311}, 2022.

\bibitem[Chung et~al.(2022)Chung, Hou, Longpre, Zoph, Tay, Fedus, Li, Wang,
  Dehghani, Brahma, et~al.]{chung2022scaling}
Chung, H.~W., Hou, L., Longpre, S., Zoph, B., Tay, Y., Fedus, W., Li, E., Wang,
  X., Dehghani, M., Brahma, S., et~al.
\newblock Scaling instruction-finetuned language models.
\newblock \emph{arXiv preprint arXiv:2210.11416}, 2022.

\bibitem[Cobbe et~al.(2021)Cobbe, Kosaraju, Bavarian, Chen, Jun, Kaiser,
  Plappert, Tworek, Hilton, Nakano, et~al.]{cobbe2021training}
Cobbe, K., Kosaraju, V., Bavarian, M., Chen, M., Jun, H., Kaiser, L., Plappert,
  M., Tworek, J., Hilton, J., Nakano, R., et~al.
\newblock Training verifiers to solve math word problems.
\newblock \emph{arXiv preprint arXiv:2110.14168}, 2021.

\bibitem[Databricks(2023)]{dolly}
Databricks.
\newblock Free dolly: Introducing the world's first truly open
  instruction-tuned llm.
\newblock Blog post, April 2023.
\newblock URL
  \url{https://www.databricks.com/blog/2023/04/12/dolly-first-open-commercially-viable-instruction-tuned-llm}.

\bibitem[Fu \& Khot(2022)Fu and Khot]{fu2022gptroadmap}
Fu, Yao;~Peng, H. and Khot, T.
\newblock How does gpt obtain its ability? tracing emergent abilities of
  language models to their sources.
\newblock \emph{Yao Fu’s Notion}, Dec 2022.
\newblock URL
  \url{https://yaofu.notion.site/How-does-GPT-Obtain-its-Ability-Tracing-Emergent-Abilities-of-Language-Models-to-their-Sources-b9a57ac0fcf74f30a1ab9e3e36fa1dc1}.

\bibitem[Fu et~al.(2022)Fu, Peng, Sabharwal, Clark, and Khot]{fu2022complexity}
Fu, Y., Peng, H., Sabharwal, A., Clark, P., and Khot, T.
\newblock Complexity-based prompting for multi-step reasoning.
\newblock \emph{arXiv preprint arXiv:2210.00720}, 2022.

\bibitem[Geng et~al.(2023)Geng, Gudibande, Liu, Wallace, Abbeel, Levine, and
  Song]{koala_blogpost_2023}
Geng, X., Gudibande, A., Liu, H., Wallace, E., Abbeel, P., Levine, S., and
  Song, D.
\newblock Koala: A dialogue model for academic research.
\newblock Blog post, April 2023.
\newblock URL \url{https://bair.berkeley.edu/blog/2023/04/03/koala/}.

\bibitem[Glaese et~al.(2022)Glaese, McAleese, Trebacz, Aslanides, Firoiu,
  Ewalds, Rauh, Weidinger, Chadwick, Thacker, et~al.]{glaese2022improving}
Glaese, A., McAleese, N., Trebacz, M., Aslanides, J., Firoiu, V., Ewalds, T.,
  Rauh, M., Weidinger, L., Chadwick, M., Thacker, P., et~al.
\newblock Improving alignment of dialogue agents via targeted human judgements.
\newblock \emph{arXiv preprint arXiv:2209.14375}, 2022.

\bibitem[Gudibande et~al.(2023)Gudibande, Wallace, Snell, Geng, Liu, Abbeel,
  Levine, and Song]{arnav2023falsepromise}
Gudibande, A., Wallace, E., Snell, C., Geng, X., Liu, H., Abbeel, P., Levine,
  S., and Song, D.
\newblock The false promise of imitating proprietary llms.
\newblock \emph{arXiv preprint arXiv:2305.15717}, 2023.

\bibitem[Hendrycks et~al.(2020)Hendrycks, Burns, Basart, Zou, Mazeika, Song,
  and Steinhardt]{hendrycks2020measuring}
Hendrycks, D., Burns, C., Basart, S., Zou, A., Mazeika, M., Song, D., and
  Steinhardt, J.
\newblock Measuring massive multitask language understanding.
\newblock \emph{arXiv preprint arXiv:2009.03300}, 2020.

\bibitem[Hendrycks et~al.(2021)Hendrycks, Burns, Kadavath, Arora, Basart, Tang,
  Song, and Steinhardt]{hendrycksmath2021}
Hendrycks, D., Burns, C., Kadavath, S., Arora, A., Basart, S., Tang, E., Song,
  D., and Steinhardt, J.
\newblock Measuring mathematical problem solving with the math dataset.
\newblock \emph{NeurIPS}, 2021.

\bibitem[Hoffmann et~al.(2022)Hoffmann, Borgeaud, Mensch, Buchatskaya, Cai,
  Rutherford, Casas, Hendricks, Welbl, Clark, et~al.]{hoffmann2022training}
Hoffmann, J., Borgeaud, S., Mensch, A., Buchatskaya, E., Cai, T., Rutherford,
  E., Casas, D. d.~L., Hendricks, L.~A., Welbl, J., Clark, A., et~al.
\newblock Training compute-optimal large language models.
\newblock \emph{arXiv preprint arXiv:2203.15556}, 2022.

\bibitem[Huang et~al.(2023)Huang, Bai, Zhu, Zhang, Zhang, Su, Liu, Lv, Zhang,
  Lei, et~al.]{huang2023c}
Huang, Y., Bai, Y., Zhu, Z., Zhang, J., Zhang, J., Su, T., Liu, J., Lv, C.,
  Zhang, Y., Lei, J., et~al.
\newblock C-eval: A multi-level multi-discipline chinese evaluation suite for
  foundation models.
\newblock \emph{arXiv preprint arXiv:2305.08322}, 2023.

\bibitem[Lewkowycz et~al.(2022)Lewkowycz, Andreassen, Dohan, Dyer, Michalewski,
  Ramasesh, Slone, Anil, Schlag, Gutman-Solo, et~al.]{lewkowycz2022solving}
Lewkowycz, A., Andreassen, A., Dohan, D., Dyer, E., Michalewski, H., Ramasesh,
  V., Slone, A., Anil, C., Schlag, I., Gutman-Solo, T., et~al.
\newblock Solving quantitative reasoning problems with language models.
\newblock \emph{arXiv preprint arXiv:2206.14858}, 2022.

\bibitem[Liang et~al.(2022)Liang, Bommasani, Lee, Tsipras, Soylu, Yasunaga,
  Zhang, Narayanan, Wu, Kumar, et~al.]{liang2022holistic}
Liang, P., Bommasani, R., Lee, T., Tsipras, D., Soylu, D., Yasunaga, M., Zhang,
  Y., Narayanan, D., Wu, Y., Kumar, A., et~al.
\newblock Holistic evaluation of language models.
\newblock \emph{arXiv preprint arXiv:2211.09110}, 2022.

\bibitem[OpenAI(2023{\natexlab{a}})]{gpt-4}
OpenAI.
\newblock Gpt-4 technical report.
\newblock \emph{arXiv preprint arXiv:2303.08774}, 2023{\natexlab{a}}.

\bibitem[OpenAI(2023{\natexlab{b}})]{openai-blog-gpt4}
OpenAI.
\newblock Gpt-4, 2023{\natexlab{b}}.
\newblock URL \url{https://openai.com/research/gpt-4}.

\bibitem[Suzgun et~al.(2022)Suzgun, Scales, Sch{\"a}rli, Gehrmann, Tay, Chung,
  Chowdhery, Le, Chi, Zhou, et~al.]{suzgun2022challenging}
Suzgun, M., Scales, N., Sch{\"a}rli, N., Gehrmann, S., Tay, Y., Chung, H.~W.,
  Chowdhery, A., Le, Q.~V., Chi, E.~H., Zhou, D., et~al.
\newblock Challenging big-bench tasks and whether chain-of-thought can solve
  them.
\newblock \emph{arXiv preprint arXiv:2210.09261}, 2022.

\bibitem[Taori et~al.(2023)Taori, Gulrajani, Zhang, Dubois, Li, Guestrin,
  Liang, and Hashimoto]{alpaca}
Taori, R., Gulrajani, I., Zhang, T., Dubois, Y., Li, X., Guestrin, C., Liang,
  P., and Hashimoto, T.~B.
\newblock Stanford alpaca: An instruction-following llama model.
\newblock \url{https://github.com/tatsu-lab/stanford_alpaca}, 2023.

\bibitem[Touvron et~al.(2023)Touvron, Lavril, Izacard, Martinet, Lachaux,
  Lacroix, Rozi{\`e}re, Goyal, Hambro, Azhar, et~al.]{touvron2023llama}
Touvron, H., Lavril, T., Izacard, G., Martinet, X., Lachaux, M.-A., Lacroix,
  T., Rozi{\`e}re, B., Goyal, N., Hambro, E., Azhar, F., et~al.
\newblock Llama: Open and efficient foundation language models.
\newblock \emph{arXiv preprint arXiv:2302.13971}, 2023.

\bibitem[Wei et~al.(2022{\natexlab{a}})Wei, Tay, Bommasani, Raffel, Zoph,
  Borgeaud, Yogatama, Bosma, Zhou, Metzler, et~al.]{wei2022emergent}
Wei, J., Tay, Y., Bommasani, R., Raffel, C., Zoph, B., Borgeaud, S., Yogatama,
  D., Bosma, M., Zhou, D., Metzler, D., et~al.
\newblock Emergent abilities of large language models.
\newblock \emph{arXiv preprint arXiv:2206.07682}, 2022{\natexlab{a}}.

\bibitem[Wei et~al.(2022{\natexlab{b}})Wei, Wang, Schuurmans, Bosma, Chi, Le,
  and Zhou]{wei2022chain}
Wei, J., Wang, X., Schuurmans, D., Bosma, M., Chi, E., Le, Q., and Zhou, D.
\newblock Chain of thought prompting elicits reasoning in large language
  models.
\newblock \emph{arXiv preprint arXiv:2201.11903}, 2022{\natexlab{b}}.

\end{thebibliography}
\bibliographystyle{icml2023}



\end{document}